# Transfer learning to enhance amenorrhea status prediction in cancer and fertility data with missing values

Xuetong Wu, Hadi Akbarzadeh Khorshidi, Uwe Aickelin, Zobaida Edib and Michelle Peate

University of Melbourne
Victoria, Australia
Email: hadi.khorshidi@unimelb.edu.au

## Contents

## 1  Introduction

Collecting sufficient labelled training data for health and medical problems is difficult (Antropova, et al., 2018). Also, missing values are unavoidable in health and medical datasets and tackling the problem arising from the inadequate instances and missingness is not straightforward (Snell, et al. 2017, Sterne, et al.

2009). However, machine learning algorithms have achieved significant success in many real-world healthcare problems, such as regression and classification and these techniques could possibly be a way to resolve the issues.

Amenorrhea status (i.e. a marker for infertility) prediction post-cancer treatment is crucial for women who wish to conceive in the future as this can guide fertility preservation decisions before they receive infertility-causing cancer treatment and post-treatment contraceptive choices (Peate, Meiser, et al. 2011). However, collecting substantial labelled data for amenorrhea prediction after cancer is challenging and very often the relevant data will present vast amount of missing values (Peate & Edib 2019).

Traditional machine learning algorithms start with the hypothesis that the training dataset and testing dataset have the same input space and distribution, which may not be practical in the real world. To address this issue, constructing a general learning model which can adapt to several similar domains quickly is necessary. Such a framework will reduce the cost of re-building and re-calibrating the learning models due to changes of distribution and input space features, which is known as *'transfer learning'*. Transfer learning is useful in many real world applications, such as natural language processing(NLP) (Han & Eisenstein 2019, Kim, Gao & Ney 2019), medical and clinical analysis (Christodoulidis, et al. 2016, Uran, et al. 2019), E-commerce (Zhao, Li, Shuai, & Yang, 2018) and acoustic recognition (Gharib, Drossos, Çakir, Serdyuk, & Virtanen, 2018). In our study, we aim to address the distribution divergence in the a large dataset consisting of several subsets by transfer learning. We firstly define the objective subset as *target domain* and impute missing values to align the feature spaces with another auxiliary subset, namely the *source domain*. Second, we try to improve classification accuracy by leveraging the information across two subsets where a single target dataset may not be sufficiently expressed.

The prepared dataset describes the relationship between the health status and amenorrhea status at 12 months from the start of the chemotherapy, for breast cancer patients internationally. The dataset includes six sub-datasets (the names of these datasets are masked due to confidentiality issues) and these sub-datasets were collected from different institutions/hospitals. Consequently, the data is not likely to be sampled from exactly the same distribution but shares some similarity, e.g., there are many common features between two subsets. To better predict the amenorrhea status across different subsets, we need to first align features using cross imputation, then map the source and target to a common latent space to maximise correlation.

The flowchart of our method is shown in Figure 1, where the learning process is divided into training and testing phases. As transfer learning aims to maximise the correlation between the source and target, our imputation and classification methods are both distance-based. In the training phase, we use zero or $k$−nearest neighbour imputation ($kNNI$) to align the features and we assume the source dataset will be more abundant while target instances are more limited. Then we perform closest pairing considering the label before canonical correlation analysis (CCA), mapping vectors $w_T$ and $w_S$ are then learned to transform paired source and target to a common latent space, eventually distance-based classifier such as $k$−nearest neighbour ($kNN$) is learned in this common space. The contributions of the paper are summarized as follows.

    1. The work explores the impact of imputation techniques on clinical datasets and provides an efficient way to deal with missing values for transfer learning.

    2. As a single dataset is insufficiently expressed, we leverage another similar dataset and further improve the performance using linear CCA, kernel CCA and deep CCA regarding the classification tasks on prediction of amenorrhoea 12 months after breast cancer diagnosis and treatment. The results show that the transfer with kernel CCA and deep CCA can boost the classification accuracy and yield promising improvements.

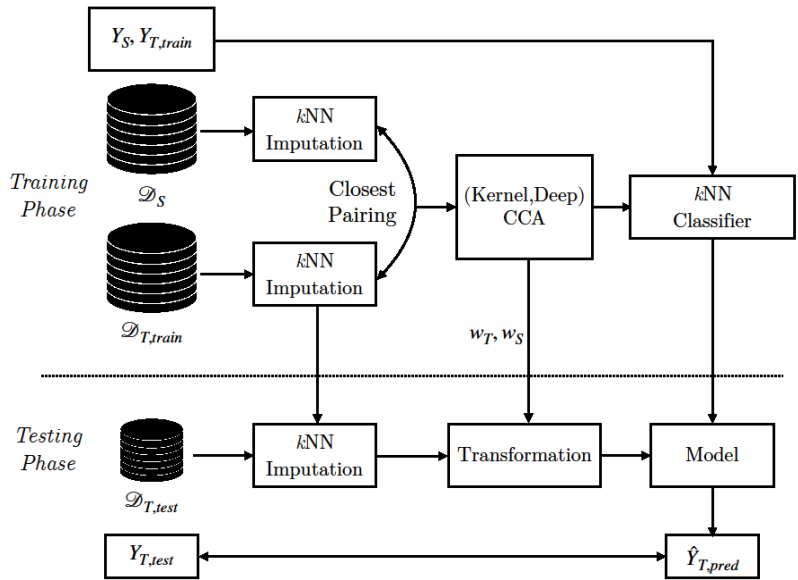

*Figure 1:* Supervised Pairing CCA method

The paper is structured as follows: section 2 introduces related studies on amenorrhea prediction to cancer status, missing values with imputation techniques and CCA-based transfer learning methods. Section 3 proposes our methods in detail. The experiments are conducted and the results are portrayed and discussed in section 4. Section 5 concludes the paper.

## 2 Related Work

### 2.1 Amenorrhea status prediction

Chemotherapy-related amenorrhea (CRA) is usually caused by gonadotoxic chemotherapy, younger (pre-menopausal) patients should be informed of the possibility of amenorrhea or recovery of menstruation and contraceptive choices (Peate, Meiser, et al. 2011) to plan for further pregnancy. Lee, et al. (2009) pointed out that the occurrence of CRA is predicted by the age at diagnosis. For those who are older than 40 years, CRA is more likely to occur and be permanent. Also Liem, et al. (2015) reported that the age at diagnosis is the main factor correlated to post-cancer infertility. Apart from age, CRA also hinges on personal factors (REF). Peate, Meiser, et al. (2011) found that low knowledge can reduce the quality of decision making. To conclude, prediction of chemotherapy-related infertility involves consideration of complex factors such as age, lifestyle factors, fertility history, ovulation, history of previous medical and gynaecological diseases, cancer-related factors and type of treatment. (Johnson, et al. 2006). Decision support is critical in ensuring patients can make informed choices about fertility preservation in a timely manner, but in practice women are making this decision without knowing their infertility risk, which has the potential for adverse effects. The key challenge with fertility prediction is that the data we use usually contains substantial missing elements which adversely impact the prediction results. Imputation methods can help to accommodate this issue.

### 2.2 Missing Value Imputation

Missing values are unavoidable in clinical datasets and this lack of information has serious drawback for data analysis. The reasons for missing data may differ, relevant knowledge cannot be acquired promptly, data will be absent due to unpredictable factors, or the cost for accessing the data is unaffordably high. Types of missing data are defined by Little and Rubin (Little & Rubin, 2019), who categorizes missing data into three types, which are *missing completely at random* (MCAR), *missing at random* (MAR) and *missing not at random* (MNAR).

MCAR cases happen when the missingness is independent of the variable itself or any other related factors, for example, chemical data may be lost accidentally, some occasional collection is omitted for questionnaires, or a few medical records will present manual documenting errors. MAR is the case when the missing representation is independent of the variable itself but can be predicted from the observed entries. A typical case is that young breast cancer patients have more missingness in fertility, compared with older patients, which can be shown by leveraging the observed age information. MNAR situations occur when the missingness is related to the variable itself, and this type of missing data cannot be predicted only from the present data. For example, breast cancer patients will be more inclined to conceal private information unrelated to the cancer such as education and salary levels, which are unlikely to be predictable. Handing this category of missing data is problematic and there are no general methods that can resolve this issue properly.

In our case, as the MNAR type is rare in the mixture of different missing data types (Goeij, et al., 2013), we may consider that the missing values are only under MCAR or MAR assumptions if a feature is not totally missing. When missing data are MCAR or MAR, they are termed 'ignorable' or 'learnable', which implies that researchers can impute data with certain procedures, e.g., by statistical analysis or machine learning approaches. Machine learning has achieved great success in many fields and the flexibility allows us to capture high-order interactions in the data (Jerez, et al., 2010) and thus impute missing values.

Let us quickly revisit the $kNNI$ routines related machine learning concepts. $kNNI$ is a type of hot deck supervised learning method, providing a path to find the most similar cases for given instances, in which nearest neighbour is a useful algorithm that matches a case with its closest $k$ neighbours in the multi-dimensional space. For missing data imputation, $kNNI$ aims to find the nearest neighbours to minimize the heterogeneous euclidean-overlap metric distance (Wilson & Martinez, 1997) between two samples, and missing items are further substituted with the values from $k$ complete cases. The advantage of $kNNI$ is that it is a simple and comprehensive method and it is suitable for large amount of missing data, but the disadvantage is that it has high computational complexity as it will compare all datasets and find the most similar cases.

### 2.3 Transfer Learning

Transfer learning resolves the issues that the training source and testing target are drawn from different distribution where a common classifier usually does not perform well. Formally, transfer learning is defined as follows (Pan & Yang 2009).

**Definition 1** *(Transfer learning) Given two different domains, namely, source domain and target domain. Given a source domain $\mathcal{D}_S$ and the source task $\mathcal{T}_S$, a target domain $\mathcal{D}_T$ and the target task $\mathcal{T}_T$, transfer learning aims to help improve the learning of the target task $\mathcal{T}_T$ using the knowledge in both source and target domains, where $\mathcal{D}_S \neq \mathcal{D}_T$, or $\mathcal{T}_S \neq \mathcal{T}_T$.*

In the definition above, the domains are not equal that $\mathcal{D}_S \neq \mathcal{D}_T$ implies that either $\mathcal{X}_S \neq \mathcal{X}_T$ or $P_S(X) \neq P_T(X)$. Similarly, $\mathcal{T}_S \neq \mathcal{T}_T$ implies that $\mathcal{Y}_S \neq \mathcal{Y}_T$ or $P_S(y|X) \neq P_T(y|X)$. In traditional machine learning methods, $\mathcal{D}_S = \mathcal{D}_T$ and $\mathcal{T}_S = \mathcal{T}_T$. Weiss, Khoshgoftaar and Wang (2016) categorize the transfer learning into homogeneous and heterogeneous types. Homogeneous transfer learning (HomoTL) assumes that the input instances are drawn from the same input space and distribution in both source and target but the tasks are different. In heterogeneous transfer learning (HeteTL) , features from source and target do not share the same feature space($\mathcal{X}_S \neq \mathcal{X}_T$), a typical case is transferring the info from image to text (Zhao, Sun, Hong, Yao, & Wang, 2019) and it does not require that the inputs should have an identical space and distribution. Formally, homogeneous and heterogeneous transfer learning are defined as follows.

**Definition 2** *(Homogeneous Transfer Learning) Given a source domain $\mathcal{D}_S$ and the source task $\mathcal{T}_S$, a target domain $\mathcal{D}_T$ and the target task $\mathcal{T}_T$, Homogeneous* *transfer learning is a type of transfer learning, where $\mathcal{D}_S = \mathcal{D}_T$, but $\mathcal{T}_S \neq \mathcal{T}_T$.*

**Definition 3** *(Heterogeneous Transfer Learning) Given a source domain $\mathcal{D}_S$ and the source task $\mathcal{T}_S$, a target domain $\mathcal{D}_T$ and the target task $\mathcal{T}_T$, Heterogeneous transfer learning is a type of transfer learning, where $\mathcal{D}_S \neq \mathcal{D}_T$, but $\mathcal{T}_S$ can be either equal or not equal to the target task $\mathcal{T}_T$.*

More specifically, the transfer learning hierarchy (Long, 2014) is illustrated in Figure 2. In our study, the prepared data consists of different sub-datasets, which have different distributions and input spaces, implying that $\mathcal{D}_S \neq \mathcal{D}_T$. To tackle this problem, we will first appropriately transform the heterogeneous transfer learning problem into homogeneous learning problem and then utilize existing homogeneous learning methods (in the domain adapatation category) for better classification.

### *2.3.1 CCA-based Transfer Learning*

Canonical correlation analysis (CCA) is a correlation-based multivariate data analysis tool that finds the maximum correlation between two sets of features in a specific subspace. In other words, CCA aims to learn a transformation that projects two datasets into a common subspace where transformed features are maximally correlated. CCA has been widely used in many fields such as software defect prediction (Jing, Wu, Dong, Qi, & Xu, 2015), computer vision and natural language processing (Hardoon, Szedmak and Shawe-Taylor 2004), medical (Parkhomenko, Tritchler and Beyene 2009) and acoustic processing (Sargin, Yemez, Erzin, & Tekalp, 2007). CCA is used for dimensionality reduction, feature mapping and learning and fusing multiple modalities for prediction. CCA methods require that both datasets are paired, that is, the number of instances should be identical. As the optimization problem is non-linear, we can map the datasets to the reproducing kernel Hilbert space (RKHS), leading to the kernel version of CCA (KCCA) (Fukumizu, Bach, & Gretton, 2007). To accurately model complex datasets with mixed types, deep neural networks are introduced to learn the corresponding representations, referring to deep CCA (DCCA) (Andrew, Arora, Bilmes, & Livescu, 2013). In this section, we will briefly revisit these three types of CCA.

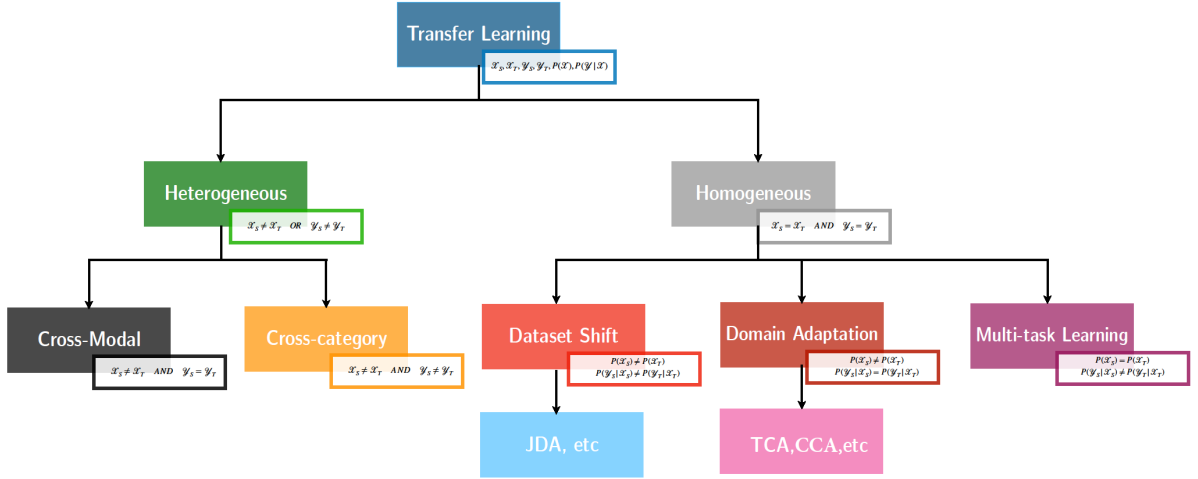

*Figure 2:* Types of transfer learning based on the feature space, label space, feature marginal distribution and conditional distribution.

**Standard Linear CCA with regularization**

Given paired source and target datasets $\hat{X}_S \in \mathbb{R}^{\mathcal{F}_U \times N}, \hat{X}_T \in \mathbb{R}^{\mathcal{F}_U \times N}$, CCA finds pairs of linear projections $w_S \in \mathbb{R}^{\mathcal{F}_U \times r}, w_T \in \mathbb{R}^{\mathcal{F}_U \times r}$ of the two views ($w_S^T \hat{X}_S \in \mathbb{R}^{r \times N}, w_T^T \hat{X}_T \in \mathbb{R}^{r \times N}$) that are maximally correlated, where $(\cdot)^T$ denotes the transpose. The optimization problem can be described as follows

$$(w_S^*, w_T^*) = \arg\max_{w_S, w_T} corr(w_S^T \hat{X}_S, w_T^T \hat{X}_T)$$

$$= \arg\max_{w_S, w_T} \frac{w_S^T cov(\hat{X}_S \hat{X}_T^T) w_T}{\sqrt{w_S^T var(\hat{X}_S, \hat{X}_S^T) w_S w_T^T var(\hat{X}_T, \hat{X}_T^T) w_T}} \quad (1)$$

Where the cross-covariance $cov(w_S^T \hat{X}_S, w_T^T \hat{X}_T)$ and variance $var(\hat{X}_S, \hat{X}_S^T)$ and $var(\hat{X}_T, \hat{X}_T^T)$ are defined as

$$var(\hat{X}_S, \hat{X}_S^T) := \frac{1}{N} \sum_{i=1}^{N} (\hat{X}_S^i - \mu_S)(\hat{X}_S^i - \mu_S)^T \in \mathbb{R}^{\mathcal{F}_U \times \mathcal{F}_U}$$

$$var(\hat{X}_T, \hat{X}_T^T) := \frac{1}{N} \sum_{i=1}^{N} (\hat{X}_T^i - \mu_T)(\hat{X}_T^i - \mu_T)^T \in \mathbb{R}^{\mathcal{F}_U \times \mathcal{F}_U}$$

$$cov(\hat{X}_S, \hat{X}_T) := \frac{1}{N} \sum_{j=1}^{N} (\hat{X}_S^i - \mu_S)(\hat{X}_T^i - \mu_T)^T \in \mathbb{R}^{\mathcal{F}_U \times \mathcal{F}_U}$$

where $\hat{X}_S^i \in \mathbb{R}_U^{\mathcal{F}}$, for $i = 1,2,\ldots,N$ and $\hat{X}_T^i \in \mathbb{R}^{\mathcal{F}_U}$, for $i = 1,2,\ldots,M$ are the instances from the unified source and target dataset. And $\mu_S \in \mathbb{R}^{\mathcal{F}_U}, \mu_T \in \mathbb{R}^{\mathcal{F}_U}$ denote the mean vector for unified source and target datasets as

$$\mu_S := \frac{1}{N} \sum_{i=1}^{N} \hat{X}_S^i$$

$$\mu_T := \frac{1}{N} \sum_{i=1}^{N} \hat{X}_T^i$$

By normalization, CCA finds the maximum canonical correlation as:

$$\text{Maximise} \quad w_S^T \Sigma_{ST} w_T$$
$$\text{Subject to} \quad w_S^T \Sigma_S w_S = 1, w_T^T \Sigma_T w_T = 1$$

(2)

The solution to this optimization problem is given by
$$\Sigma_{ST}\Sigma_T^{-1}\Sigma_{TS}\mathbf{w}_S = \lambda^2 \Sigma_S \mathbf{w}_S$$
$$\Sigma_S^{-1}\Sigma_{ST}\Sigma_T^{-1}\Sigma_{TS}\mathbf{w}_S = \lambda^2 \mathbf{w}_S$$

which can also be written as
$$\begin{pmatrix} 0 & \Sigma_{ST} \\ \Sigma_{TS} & 0 \end{pmatrix} \begin{pmatrix} w_S \\ w_T \end{pmatrix} = \lambda^2 \begin{pmatrix} \Sigma_S & 0 \\ 0 & \Sigma_T \end{pmatrix} \begin{pmatrix} w_S \\ w_T \end{pmatrix} \quad (3)$$

Equation (3) leaves an eigenvalue problems. Regularized CCA is introduced to address the problem that if $\Sigma_S$ and $\Sigma_T$ are singular then CCA is ill-posed and the generalized eigenvalue problem cannot be solved properly. Imposing the L2 penalty maintains the convexity of the problem and the generalized formulation. The optimization object function is expressed by

$$(w_S^*, w_T^*) = \arg \max_{w_S, w_T} corr(w_S^T \hat{X}_S, w_T^T \hat{X}_T)$$

$$= \arg \max_{w_S, w_T} \frac{w_S^T cov(\hat{X}_S \hat{X}_T^T) w_T}{\sqrt{(w_S^T var(\hat{X}_S, \hat{X}_S^T) w_S + \rho ||w_S||^2)(w_T^T var(\hat{X}_T, \hat{X}_T^T) w_T + \rho ||w_T||^2)}}$$

(4)

The eigenvalue problem is formulated by
$$\begin{pmatrix} 0 & \Sigma_{ST} \\ \Sigma_{TS} & 0 \end{pmatrix} \begin{pmatrix} w_S \\ w_T \end{pmatrix} = \lambda^2 \begin{pmatrix} \Sigma_S + \rho I & 0 \\ 0 & \Sigma_T + \rho I \end{pmatrix} \begin{pmatrix} w_S \\ w_T \end{pmatrix}$$

and the solution to the problem above is to find the largest $r$ eigenvalues for the matrix

$$\begin{pmatrix} \Sigma_S + \rho I & 0 \\ 0 & \Sigma_T + \rho I \end{pmatrix}^{-1} \begin{pmatrix} 0 & \Sigma_{ST} \\ \Sigma_{TS} & 0 \end{pmatrix}$$

### Kernel CCA

KCCA finds a pair of nonlinear projection of the two views. The functions in RKHS are denoted as $\mathcal{H}_S, \mathcal{H}_T$ and the associated positive definite kernels are denoted as $\Phi_S, \Phi_T$. The optimal projections from low dimension $\mathcal{F}_U$ to high dimension $\mathcal{F}_U^H$ are any functions $h_S \in \mathcal{H}_S, h_T \in \mathcal{H}_T$ to maxise the correlation as

$$(h_S^*, h_T^*) = \underset{f_S \in \mathcal{H}_S, f_T \in \mathcal{H}_T}{\arg\max} corr\left(h_S(\hat{X}_S), h_T(\hat{X}_T)\right)$$

$$= \underset{h_S \in \mathcal{H}_S, h_T \in \mathcal{H}_T}{\arg\max} \frac{cov(h_S(\hat{X}_S), h_T(\hat{X}_T))}{\sqrt{var(h_S(\hat{X}_S))var(h_T(\hat{X}_T))}}$$

$$= \frac{w_S K_S K_T w_T}{\sqrt{w_S^T K_S^2 w_S}\sqrt{w_T^T K_T^2 w_T}} \quad (5)$$

where $w_S, w_T \in \mathbb{R}^{\mathcal{F}_U^H}$, and we define the centralized kernel matrix $K_S, K_T \in \mathbb{R}^{\mathcal{F}_U^H \times \mathcal{F}_U^H}$ as

$$K_S := \frac{1}{N} \sum_{j=1}^{N} \left(\Phi(\hat{x}_S^j) - \hat{u}_S\right)\left(\Phi(\hat{x}_S^j) - \hat{u}_S\right)^T$$

$$K_T := \frac{1}{N} \sum_{j=1}^{N} \left( \Phi(\hat{x}_T^j) - \hat{u}_T \right)\left( \Phi(\hat{x}_T^j) - \hat{u}_T \right)^T$$

where $\Phi(\hat{x}_S^j) \in \mathbb{R}^{\mathcal{F}_U^H}$, for $j = 1,2,\cdots, N$, and $\Phi(\hat{x}_T^j) \in \mathbb{R}^{\mathcal{F}_U^H}$, for $j = 1,2,\cdots, N$, $\hat{u}_S$ and $\hat{u}_T$ are the mean values for transformed data matrix regarding each feature. By normalization, KCCA finds the maximum canonical correlation with

$$\text{Maximise} \quad w_S^T K_S K_T w_T$$
$$\text{Subject to} \quad w_S^T K_S^2 w_S = 1, w_T^T K_T^2 w_T = 1 \quad (6)$$

which can be modified to the following generalized eigenvalue problems:

$$\begin{bmatrix} 0 & K_S K_T \\ K_T K_S & 0 \end{bmatrix} \begin{bmatrix} w_S \\ w_T \end{bmatrix} = \lambda \begin{bmatrix} K_S^2 & 0 \\ 0 & K_T^2 \end{bmatrix} \begin{bmatrix} w_S \\ w_T \end{bmatrix}$$

### Deep CCA

Instead of constructing standard CCA, deep CCA using two neural networks is illustrated as follows:

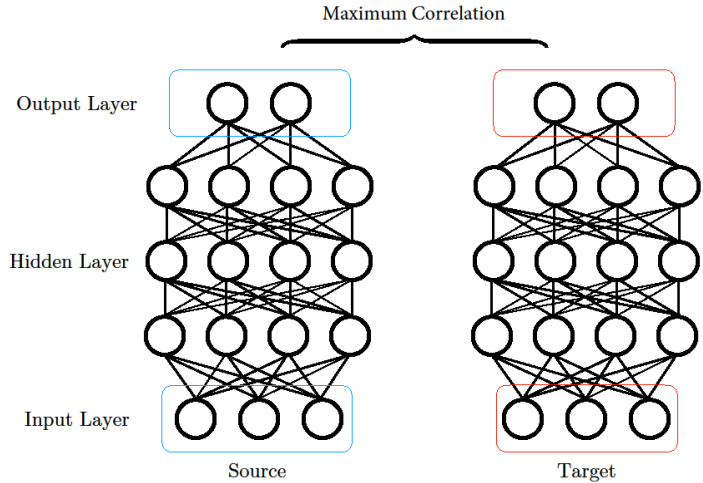

*Figure 3: Deep CCA network structure, blue nodes correspond to input source samples, red nodes are target samples, and output layers of source and target are maximised.*

The deep neural network is used to learn a common latent space where the correlation between two views are as high as possible. The neural network shown in Figure 3 are defined as $w_S$ and $w_T$ in both views. We denote the neural network models of source and target views as $f_S(\cdot)$ and $f_T(\cdot)$, then we aim to find the optimal $w_S^*$ and $w_T^*$ where

$$(w_S^*, w_T^*) = \underset{(w_S, w_T)}{argmax}\, corr\left( f_S(\hat{X}_S; w_S), f_T(\hat{X}_T; w_T) \right) \quad (7)$$

Let $H_S \in \mathbb{R}^{r \times N}$ and $H_T \in \mathbb{R}^{r \times N}$ be the learned representations produced by the deep models on the two views. To make the matrices centered, we define $\overline{H}_S = H_S - \frac{1}{N} H_S \mathbf{I}_N$, and the sample co-variance is defined as $\hat{\Sigma}_{ST} = \frac{1}{N} \overline{H}_S \overline{H}_T^T$, and the variance for source domain is given by $\hat{\Sigma}_S = \frac{1}{N} \overline{H}_S \overline{H}_S^T + \lambda_S I$ with regularization constant $\lambda_S$. The centered matrix and variance for target domain is defined similarly. The total correlation of the top $r$ components of $H_S$ and $H_T$ is the sum of the top $r$ singular values of the matrix $\hat{\Sigma}_r = \hat{\Sigma}_S^{-1/2} \hat{\Sigma}_{ST} \hat{\Sigma}_T^{-1/2}$. Using the singular value decomposition of $\hat{\Sigma}_r = UDV^T$, then the gradient for the source data is computed by

$$\frac{\partial corr(H_S, H_T)}{\partial H_S} = \frac{1}{N}\left(2\nabla_S \overline{H}_S + \nabla_{ST} \overline{H}_T\right)$$

where

$$\nabla_{ST} = \hat{\Sigma}_S^{-1/2} UV^T \hat{\Sigma}_T^{-1/2}$$
$$\nabla_S = -\frac{1}{2} \hat{\Sigma}_S^{-1/2} UDU^T \hat{\Sigma}_S^{-1/2}$$

The gradient for the target domain is symmetric

$$\frac{\partial corr(H_S, H_T)}{\partial H_T} = \frac{1}{N}\left(2\nabla_T \overline{H}_T + \nabla_{ST} \overline{H}_S\right)$$

where

$$\nabla_{ST} = \hat{\Sigma}_S^{-1/2} UV^T \hat{\Sigma}_T^{-1/2}$$
$$\nabla_T = -\frac{1}{2} \hat{\Sigma}_T^{-1/2} UDU^T \hat{\Sigma}_T^{-1/2}$$

These gradients are back-propagated for training the neural network model.

### 3 Method
### 3.1 Unified Feature Representation with imputation

In this section, we present our method with subspace embedding diagram as shown in Figure 4, where source and target datasets are projected in a common latent space and a generic classifier is learned to distinguish the labels. Given source and target datasets $\mathcal{D}_S$ and $\mathcal{D}_T$ with $N$ and $M$ ($N \neq M$) samples, to apply CCA in both datasets, the number of instances should be identical, which is not likely for real-world scenarios. We propose a method for pairing, which is finding the nearest pair for the dataset with the smaller number of instances (target) in another one (source) by aligning the feature space first.

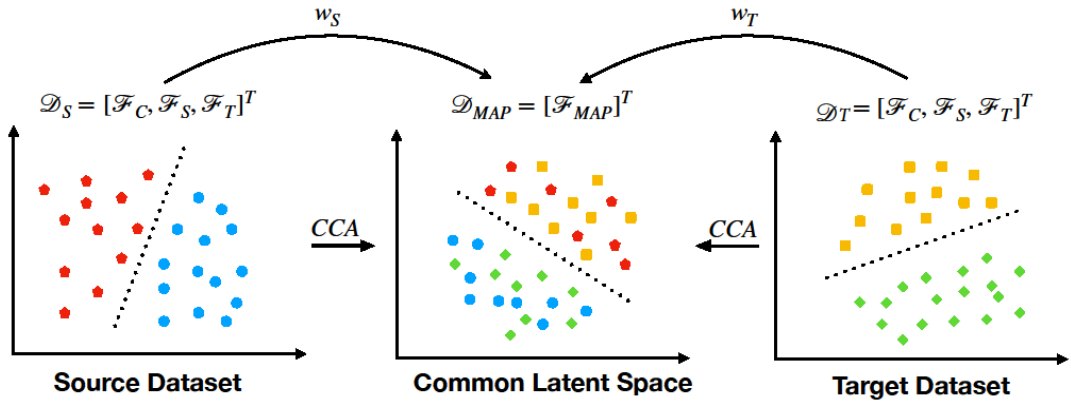

*Figure 4: Subspace Embedding Learning Representation*

Assume we have two datasets with common binary label $\mathcal{Y}$, namely, domain source and target source, which are denoted as $\mathcal{D}_S$ and $\mathcal{D}_T$, respectively. As the two databases are not fully aligned, we split the feature space into three components according to the missing rate, e.g. *common features* $F_C$, *target-specific features* $F_T$ and *source-specific features* $F_S$ as shown in the Figure 5, where $F_C$ are the features that both $\mathcal{D}_S$ and $\mathcal{D}_T$ observe, here the term 'observe' entails that the observed instances are substantial enough for each feature (at least one observation). $F_T$ contains features that are only observed in the target and is totally missing in the source and similarly, $F_S$ consists of features observed in the source and is totally missing in the target.

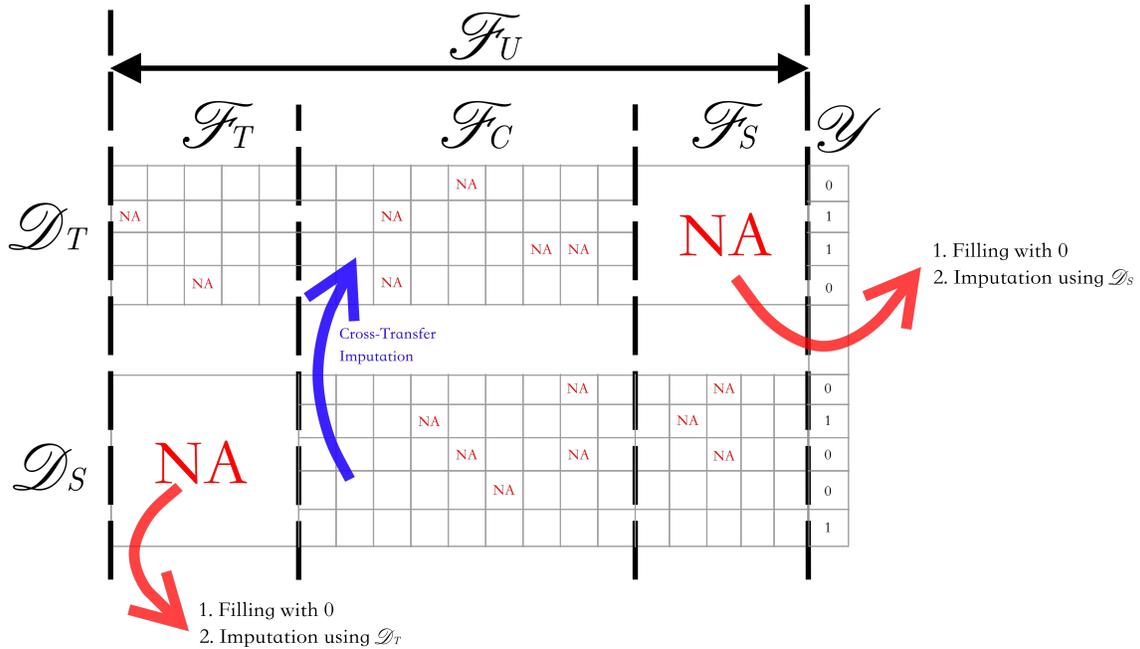

**Figure 5:** *Unified feature representation for two datasets $\mathcal{D}_T$ and $\mathcal{D}_S$. NAs in grid suggest data are missing, observed feature entries are left blank, two large NA indicate that $F_T$ and $F_S$ in $\mathcal{D}_S$ and $\mathcal{D}_T$ are totally missing. Cross-transfer and feature creation imputation are performed on source and target specific features.*

More formally, given two datasets, namely source domain $\mathcal{D}_S$ and target domain $\mathcal{D}_T$ regarding the same binary label $\mathcal{Y} \in \{0,1\}$. Input feature spaces for each domain are denoted as $\mathcal{X}_S$ and $\mathcal{X}_T$, then we define common feature space $F_C = \mathcal{X}_S \cap \mathcal{X}_T$, target-specific feature space $F_T = \mathcal{X}_T \backslash F_C$ and source-specific feature space $F_S = \mathcal{X}_S \backslash F_C$.

The imputation involves two procedures, which are *cross-transfer imputation* and *feature creation imputation*. Cross-transfer imputation aims to fill the missing values in $F_C$ using existing imputation techniques, while feature creation imputation can be achieved in two ways, one is to complete $F_T$ for the source set and $F_S$ for the target set by utilizing the data from target and source respectively and the second is to simply pad zeros for missing values in $F_T$ and $F_S$.

## 3.2 Nearest Pairing Methods

For each sample in the target dataset, we desire to identify the most similar one from the source dataset. We propose a heuristic pairing algorithm in this section. Firstly the distance matrix ($N \times M$) is created using Euclidean metrics. Each column represents source indices between all source instances and one specific target instance sorted by distance in ascending order. That is, the first row represents the target index where target instance and source instance have minimal distance and vice versa. Then in the first row of the distance matrix, we find the largest number of non-replicated instances from the source and then delete these paired instances from source and target until all pairings are finished. The algorithm is shown below.

---

**Algorithm 1:** Nearest Pairing

**Input**: Source with $N$ instances, target $M$ instances, assume $M \leq N$.

**Output:** Paired source and target datasets.

**Step-1:** Data normalization and pre-processing;

**Step-2: while** $Instance(\mathcal{D}_S) > 0$ **do**

{
  $Dist\_Mtx \leftarrow Euclidean\_Distance(\mathcal{D}_S, \mathcal{D}_T)$;
  Find largest non-replicated pairs in the first row, which are denoted as $\mathcal{D}_{S,P}, \mathcal{D}_{T,P}$;
  $\mathcal{D}_S \leftarrow \mathcal{D}_S \backslash \mathcal{D}_{S,P}$;
  $\mathcal{D}_T \leftarrow \mathcal{D}_T \backslash \mathcal{D}_{T,P}$;
}

**Step-3:** Return $\mathcal{D}_S, \mathcal{D}_T$.

---

## 3.3 CCA Transformation

> **Algorithm 2:** Classification with nearest pairing
>
> **Input**: Source: $\mathcal{D}_S \in \mathbb{R}^{n_S \times d_S}$, target: $\mathcal{D}_T \in \mathbb{R}^{n_T \times d_T}$, assume $n_S \leq n_T$.
>
> **Output:** Predict amenorrhea status after cancer treatments.
>
> **Step-1:** $\mathcal{F}_S \leftarrow \text{observed\_features}(\mathcal{D}_S)$;
>
> **Step-2:** $\mathcal{F}_T \leftarrow \text{observed\_features}(\mathcal{D}_S)$;
>
> **Step-3:** $\mathcal{F}_C \leftarrow \mathcal{F}_T \cap \mathcal{F}_S$;
>
> **Step-4:** Data normalization and pre-processing;
>
> **Step-5:** Cross-transfer imputation with $kNNI$;

Once we obtain the paired source and target datasets, we can apply the linear CCA with regularization (equation (4)), the kernel CCA (equation (6)) or the deep CCA (equation (7)) for finding a common latent space. A classifier is then employed in the common latent space. Details are shown in Algorithm 2.

## 4 Experiments and Result
### 4.1 Data description

The full dataset is authorized by the Fertility After Cancer Predictor (FoRECAsT) Study, University of Melbourne (Peate & Edib, 2019). The dataset summary and five samples are described in Table 1 and 2. The data contains 1565 samples in total with six sub-datasets. From the view of features, most are categorical and there are a few numerical ones. Each sub-dataset suffers from missingness more or less, from 8.6% to 23.6%. As the features are highly misaligned, the total missingness reaches an excessive rate, at 72.5%. In this section, we will conduct the nearest paring algorithms for the FoRECAsT datasets and compare the performance with some benchmarks in terms of classification accuracy.

*Table 1:* Data description for FoRECAsT dataset, the dataset is split into six subsets regarding the sources. Observed feature here implies that the feature has at least one observation within the dataset, and missingness relates to the observed features.

| Data track | Instances | Observed Features | Categorical | Numerical | Misingness | Label |
|---|---|---|---|---|---|---|
| Track 0 | 725 | 19 | 19 | 0 | 8.6% | |
| Track 1 | 280 | 36 | 36 | 0 | 9.1% | |
| Track 2 | 209 | 34 | 29 | 5 | 10.5% | 'Amen_ST12' |
| Track 3 | 154 | 20 | 19 | 1 | 22.2% | |
| Track 4 | 101 | 42 | 40 | 2 | 23.6% | |
| Track 5 | 96 | 47 | 43 | 4 | 18.3% | |

| | | | | | | | |
|---|---|---|---|---|---|---|---|
| Overall | 1565 | 87 | 76 | 11 | 72.5% | | |

Table 2: *Five instances sampled from the FoRECAsT datasets, the entry 'NA' implies that the corresponding data is missing.*

| ID | Age | Age_category | Smoking | LiveBirth | Menacheage | ER | TT | … | Amen_ST12 |
|---|---|---|---|---|---|---|---|---|---|
| 5249 | 47.3949 | 0 | 0 | 1 | 13 | 1 | 1 | … | 1 |
| 5250 | 33.4565 | 1 | 0 | 1 | 15 | 1 | NA | … | 1 |
| 5258 | 52.2081 | 0 | 1 | 1 | 1 | 1 | NA | … | 1 |
| 5259 | 52.3559 | 0 | 0 | 1 | 15 | 0 | NA | … | 1 |
| 5263 | 50.4038 | 0 | 1 | 1 | NA | 1 | NA | … | 1 |

### 4.2 Result

To investigate the effectiveness of the CCA-based methods for the infertility classification task, three approaches are adopted, which include CCA, kernel CCA and deep CCA, where the benchmark is the accuracy based on a classifier using the original sub-dataset only. All modules are implemented in Python 3.7 (Van Rossum & Drake Jr, 1995) in the operating system Mac OS 10.14.3. For hyper-parameter tuning, cross imputation uses 5NN, the latent space dimensionality is set to be fixed half of the rank $r = \frac{1}{2}\min(\text{rank}(\mathcal{D}_S); \text{rank}(\mathcal{D}_T))$, the classifier is set to be 1NN as suggested by many existing papers (Jing, et al. 2015, Wang, et al. 2017). The neural network model parameters for both source and target views are summarized in Table 3.

Table 3: *The neural network model parameters for both source and target views*

| Layers | Dimension for Target | Dimension for Source | Activation Function |
|---|---|---|---|
| 1 | 512 | 512 | Sigmoid |
| 2 | 512 | 512 | Sigmoid |
| 3 | 512 | 512 | Sigmoid |
| 4 | $r$ | $r$ | Sigmoid |

#### 4.2.1 Distance Metrics Evaluation
**Proxy-A-Distance**

Ben-David et al. (2004, 2010) propose a distance metric to evaluate the distribution divergence, which is known as '$\mathcal{A}$-distance'. Computing the $\mathcal{A}$-distance can be approximated by learning a classifier, suppose we have two datasets $\mathcal{D}_S$ and $\mathcal{D}_T$, then a classifier $h$ is learned which achieves minimum error on the binary classification problem of discriminating between points genereated by the two distributions.

To see this, suppose we have two samples $U_S$ and $U_T$ sampled from the source and target datasets with the same length $m$, define the error of a classifier $h$ on the task of discriminating between points sampled from different distributions as:

$$err(h) = \frac{1}{2m} \sum_{i=1}^{2m} |h(\mathbf{x}_i) - I_{\mathbf{x}_i \in U_S}|$$

Where $I_{\mathbf{x}_i \in U_S}$ is the binary indicator that where the sample $\mathbf{x}_i$ lies in $U_S$ or not, and the proxy-$mathcalA$-distance is defined as:

$$d_A(U_S, U_T) = 2(1 - 2\min_h err(h))$$

It is important to note that it does not provide us with a valid upper bournd on the target error, but gives some intuition on how different the source and target datasets are and also gives us some useful insights about the representations for domain adaptation.

### Maximum Mean Discrepancy

Maximum mean discrepancy is a kernel evaluation metric proposed by Borgwardt, et al. (2006), which is a relevant criterion for comparing distributions based on $\mathbb{R}$. Let $X = x_1, x_2, \cdots, x_n$ and $Y = y_1, y_2, \cdots, y_m$ be random variables with distribution $\mathcal{P}$ and $\mathcal{Q}$. Then the empirical estimate of distance between $\mathcal{P}$ and $\mathcal{Q}$ is defined as:

$$Dist(X, Y) = \| \frac{1}{n} \sum_{i=1}^{n} \phi(x_i) - \frac{1}{m} \sum_{i=1}^{m} \phi(y_i) \|_{\mathcal{H}}$$

Where $\mathcal{H}$ is a universal RKHS and $\phi(\cdot)$ is a mapping function: $\mathcal{X} \to \mathcal{H}$.

### Coral Loss Function

The coral loss function is defined in (Sun & Saenko, 2016). This metric evaluates the distance between the second-order statistics(covariances) of the source and target features:

$$l_{CORAL} = \frac{1}{4d^2} \| C_S - C_T \|_F^2$$

where $\|\cdot\|_F^2$ denotes the squared matrix Frobenius norm. And the covariance matrix of the source and target data are given by:

$$C_S = \frac{1}{n_S - 1}(D_S^T D_S - \frac{1}{n_S}(\mathbf{1}^T D_S)^T(\mathbf{1}^T D_S))$$

$$C_T = \frac{1}{n_T - 1}(D_T^T D_T - \frac{1}{n_S}(\mathbf{1}^T D_T)^T(\mathbf{1}^T D_T))$$

where $\mathbf{1}$ is a column vector with all elements equal to 1. And $d$ is the number of features. And $d$ is the number of features.

### *4.2.2 Prediction Accuracy*

Transfer learning aims to extract the similarity between the source and target datasets and, many studies use the divergence metrics to evaluate how different two datasets are and try to learn a common latent space to minimize the according loss function. We will evaluate these metrics on FoRECAsT datasets and the losses will give us a useful insight on how well CCA transfer learning works. We transfer the knowledge from the dataset that contains more samples to that wieh fewer, for example, track 5 transferring to track 1 is abbreviated as $T5 \rightarrow T1$, leading to 15 sets of experiments. The distance metrics results and prediction accuracy are shown in Table 4 and 5, each column is represented as a different baseline, given as follows.
- S→T : Source dataset transfers to the target dataset.
- Original: Using the target only.
- ZPC : Closest pairing with zero padding for feature creation imputation.
- IMC : Closest pairing with $kNNI$ for feature creation imputation.
- ZPCCA : Linear CCA based on the ZPC.
- IMCCA : Linear CCA based on the IMC.
- ZPKCCA : Kernel CCA based on the ZPC using linear kernel.
- IMKCCA : Kernel CCA based on the IMC using linear kernel.
- ZPDCCA : Deep CCA based on the ZPC.
- IMDCCA : Deep CCA based on the IMC.

*Table 4:* *Distance Metrics from different baselines*

| Methods | MMD | | | | $\mathcal{A}$-distance | | | | Coral Loss | | | |
|---|---|---|---|---|---|---|---|---|---|---|---|---|
| S→T | ZPC | IMC | ZPCCA | IMCCA | ZPC | IMC | ZPCCA | IMCCA | ZPC | IMC | ZPCCA | IMCCA |
| T4→T5 | 5.18 | 0.25 | 0.00 | 0.00 | 1.96 | 1.08 | 0.00 | 0.00 | 4.7e7 | 2.1e7 | 0.00 | 0.00 |
| T3→T5 | 4.80 | 0.02 | 0.00 | 0.00 | 1.96 | 0.17 | 0.00 | 0.00 | 77.58 | 1.164 | 0.00 | 0.00 |
| T2→T5 | 1.18 | 0.09 | 0.00 | 0.00 | 2.00 | 0.68 | 0.04 | 0.00 | 2.7e9 | 7.4e7 | 0.00 | 0.00 |
| T1→T5 | 5.10 | 4.81 | 0.00 | 0.01 | 2.00 | 1.96 | 0.08 | 0.00 | 90.78 | 90.80 | 0.00 | 0.00 |
| T0→T5 | 4.80 | 0.12 | 0.03 | 0.00 | 1.96 | 0.68 | 0.17 | 0.08 | 122.4 | 4.82 | 0.00 | 0.00 |
| T3→T4 | 5.23 | 0.17 | 0.01 | 0.00 | 1.88 | 1.13 | 0.20 | 0.04 | 1.3e8 | 4.6e7 | 0.00 | 0.00 |
| T2→T4 | 3.04 | 0.11 | 0.00 | 0.00 | 1.88 | 0.78 | 0.00 | 0.00 | 2.1e9 | 1.1e8 | 0.00 | 0.00 |
| T1→T4 | 5.23 | 0.26 | 0.01 | 0.00 | 1.88 | 1.13 | 0.20 | 0.04 | 1.1e8 | 2.8e7 | 0.00 | 0.00 |
| T0→T4 | 5.23 | 0.04 | 0.14 | 0.01 | 1.88 | 0.55 | 0.20 | 0.04 | 1.8e8 | 2.2e7 | 0.03 | 0.00 |
| T2→T3 | 1.20 | 0.04 | 0.01 | 0.01 | 2.00 | 0.73 | 0.13 | 0.10 | 3.1e9 | 2.9e8 | 0.00 | 0.00 |
| T1→T3 | 6.82 | 0.22 | 0.02 | 0.02 | 2.00 | 1.40 | 0.21 | 0.05 | 0.185 | 0.009 | 0.00 | 0.00 |
| T0→T3 | 0.86 | 0.03 | 0.05 | 0.01 | 2.00 | 0.57 | 0.83 | 0.08 | 0.003 | 0.003 | 0.00 | 0.00 |

| S→T | | | | | | | | | | | | |
|---|---|---|---|---|---|---|---|---|---|---|---|---|
| T1→T2 | 1.31 | 0.30 | 0.01 | 0.01 | 2.00 | 1.28 | 0.04 | 0.00 | 3e10 | 3.4e9 | 0.00 | 0.00 |
| T0→T2 | 1.31 | 0.00 | 0.06 | 0.01 | 2.00 | 0.93 | 0.44 | 0.21 | 4.5e9 | 2.6e7 | 0.01 | 0.00 |
| T0→T1 | 6.87 | 0.22 | 0.12 | 0.03 | 2.00 | 1.63 | 1.13 | 0.87 | 0.97 | 0.02 | 0.00 | 0.00 |

In Table 5, the negative transfers are highlighted in red while positive cases are highlighted in blue. The highest accuracy is highlighted in bold for each row. We set the result from the original data as our benchmark. Zero padding and $kNNI$ for feature creation imputation are both conducted and compared with and without CCA procedures.

*Table 5: The classification results with acc±dev with 5-fold cross-validation, negative transfers are highlighted in red and positive transfers is highlighted in blue, the largest accuracy is highlighted in bold for each row.*

| S→T | Original | ZPC | IMC | ZPCCA | IMCCA | ZPKCCA | IMKCCA | ZPDCCA | IMDCCA |
|---|---|---|---|---|---|---|---|---|---|
| T4→T5 | 0.834±0.060 | 0.834±0.060 | 0.844±0.093 | 0.750±0.070 | 0.771±0.023 | 0.875±0.041 | 0.844±0.058 | 0.865±0.053 | **0.896±0.065** |
| T3→T5 | 0.834±0.060 | 0.834±0.060 | 0.834±0.060 | 0.876±0.050 | 0.845±0.071 | 0.875±0.041 | 0.875±0.041 | **0.917±0.025** | 0.906±0.052 |
| T2→T5 | 0.834±0.060 | 0.834±0.060 | 0.740±0.044 | 0.761±0.024 | 0.771±0.040 | **0.875±0.041** | 0.751±0.035 | 0.854±0.039 | 0.823±0.027 |
| T1→T5 | 0.834±0.060 | 0.834±0.060 | 0.834±0.060 | 0.751±0.086 | 0.761±0.078 | **0.875±0.041** | **0.875±0.041** | 0.854±0.051 | 0.792±0.044 |
| T0→T5 | 0.834±0.060 | 0.834±0.060 | 0.834±0.060 | 0.824±0.050 | 0.782±0.060 | 0.875±0.041 | 0.875±0.041 | **0.948±0.001** | 0.907±0.037 |
| T3→T4 | 0.644±0.085 | 0.644±0.085 | 0.605±0.078 | 0.642±0.110 | 0.575±0.059 | 0.663±0.068 | 0.634±0.021 | **0.672±0.055** | 0.662±0.126 |
| T2→T4 | 0.644±0.085 | 0.644±0.085 | 0.584±0.102 | 0.603±0.066 | 0.526±0.124 | 0.653±0.084 | 0.554±0.049 | **0.673±0.069** | 0.663±0.068 |
| T1→T4 | 0.644±0.085 | 0.644±0.085 | 0.605±0.103 | 0.603±0.092 | 0.634±0.100 | 0.693±0.087 | **0.703±0.052** | 0.623±0.043 | 0.594±0.050 |
| T0→T4 | 0.644±0.085 | 0.644±0.085 | 0.605±0.101 | 0.574±0.043 | 0.584±0.076 | 0.663±0.068 | **0.693±0.087** | 0.613±0.041 | 0.644±0.083 |
| T2→T3 | 0.590±0.050 | 0.584±0.062 | 0.578±0.087 | 0.689±0.079 | 0.571±0.090 | 0.676±0.077 | 0.519±0.086 | 0.662±0.015 | **0.714±0.048** |
| T1→T3 | 0.590±0.050 | 0.597±0.050 | 0.629±0.073 | 0.540±0.051 | 0.585±0.092 | 0.656±0.081 | **0.702±0.078** | 0.702±0.101 | 0.598±0.080 |
| T0→T3 | 0.590±0.050 | 0.590±0.050 | 0.662±0.051 | **0.727±0.060** | 0.539±0.063 | 0.715±0.078 | 0.715±0.032 | 0.643±0.074 | 0.702±0.046 |
| T1→T2 | 0.766±0.027 | 0.761±0.029 | 0.761±0.039 | **0.785±0.015** | 0.741±0.031 | 0.770±0.055 | 0.780±0.050 | 0.650±0.065 | 0.690±0.077 |

| T0→T2 | 0.766±0.027 | 0.761±0.029 | 0.770±0.041 | 0.761±0.036 | 0.727±0.068 | 0.767±0.040 | **0.770±0.038** | 0.672±0.063 | 0.683±0.068 |
|---|---|---|---|---|---|---|---|---|---|
| T0→T1 | 0.518±0.054 | 0.500±0.054 | 0.625±0.040 | 0.629±0.056 | 0.636±0.063 | 0.661±0.049 | 0.656±0.048 | 0.770±0.035 | **0.780±0.020** |
| Average | 0.704 | 0.703 | 0.701 | 0.700 | 0.670 | **0.753** | 0.730 | 0.741 | 0.737 |

### 4.3 Discussion

From the divergence results, we can observe that after CCA, both three distances metrics drop close to zero and this implies that CCA reduces the distribution differences as $P(D_{S,CCA}) \approx P(D_{T,CCA})$. Comparing the zero padding approach with $kNNI$ for $F_T$ and $F_S$ in the source and target datasets, the latter method reduces the domain divergence due to the introduction of interference from the source domain while the former maintains the structure of the data and increases the domain divergence by importing large number of zeros. However, as the algorithm does not take the conditional distribution $P(Y_S|D_{S,CCA})$ into account, negative transfer will happen if $P(Y_S|D_{S,CCA}) \neq P(Y_T|D_{T,CCA})$, i.e., the accuracy drops after transferring, compared to the original results. For example, if the classification accuracy happens to be relatively low in the source domain, it may not be helpful in distinguishing the label in the target domain. To address this, we apply kernel CCA to provide stronger discriminant power and this method gives promising accuracy improvements.

From the results of classification accuracy, $kNNI$ for feature creation imputation can help the target domain improve classification accuracy, except for subset $T4$. For linear CCA cases, negative transfer occurs regardless of the imputation techniques, despite the domain divergence being reduced, the conditional probability may still differ, which indicates that the elementary linear dimensionality reduction transformation cannot achieve promising results. When using kernel CCA, the classification becomes linearly separable in high dimensional space. Both zero-padding and $kNNI$ yield good results. As for deep CCA, zero and cross imputation can help classify the labels, though negative transfer happens for dataset $T2$. In conclusion, zero padding can achieve better results overall, while $kNNI$ is more likely to produce the best results.

In general, whilst the standard CCA minimizes the distribution difference between the source and target datasets, it still cannot generalize the knowledge efficiently from the source domain using linear transformation, except $T1 \rightarrow T2$ and $T0 \rightarrow T3$ with zero padding, while the Kernel CCA with zero padding performs strongly with no negative transfers. Regarding the highest accuracy, cross imputation with kernel CCA and zero padding with deep CCA show remarkably positive transfers. By observing the average accuracy, compared with the benchmark using the original datasets, transfer learning with kernel and deep methods achieves promising prediction results for amenorrhea prediction accuracy improvements while standard CCA cannot tackle the problem properly.

However, the algorithm requires that source instances should be larger than target, which is a limitation because the knowledge from the source dataset is not fully utilized and reverse transferring is not possible in our algorithm. In addition, finding appropriate mapping dimension is fixed to be $\frac{1}{2}\min(\text{rank}(\mathcal{D}_S); \text{rank}(\mathcal{D}_T))$ in our case. We will investigate these issues in future work.

## 5 Conclusion

The paper proposes a CCA-based transfer learning classification for the amenorrhea status prediction of breast cancer patients using nearest pairing algorithm and missing values imputation methods. To address the domain divergence issues, CCA minimizes the domain difference by maximizing the correlation between the source and the target. Utilizing kernel or deep CCA achieves ideal results and boosts the classification performance. However, the pairing algorithm introduces limitations and may cause information loss of the source. Developing efficient learning to take advantages of all source information is left to future work. In addition, the results show that the reduction in distribution divergence (measured by distance metrics) cannot guarantee increase in accuracy. There are other factors such as conditional probabilities may cause the reduction in accuracy. More investigations on these factors and correlation between distance metrics and accuracy improvement are suggestions for the further research. This study provides a new roadmap for health researchers dealing with medical data with large amounts of missing values using the transfer learning framework.

### Acknowledgement

This work is fully funded by Melbourne Research Scholarships (MRS), Granted 19/03 and partially supported by Fertility After Cancer Predictor (FoRECAsT) Study. Michelle Peate is currently supported by an MDHS Fellowship, University of Melbourne. The FoRECAsT study is supported by the FoRECAsT Consortium and Victorian Government through a Victorian Cancer Agency (Early Career Seed Grant) awarded to Michelle Peate.